\documentclass[runningheads]{llncs}
\usepackage{graphicx}
\usepackage{amsmath,amssymb} % define this before the line numbering.
\usepackage{color}
\usepackage{multirow}
\usepackage{tabularx}
\usepackage{relsize}
\usepackage{mathtools}
\usepackage{commath}
\usepackage{makecell}
\usepackage{epsfig}
\usepackage{stackengine}
\usepackage[width=122mm,left=12mm,paperwidth=146mm,height=193mm,top=12mm,paperheight=217mm]{geometry}

\usepackage{graphicx}
\usepackage{amsmath,amssymb} % define this before the line numbering.

\usepackage{times}
\usepackage{epsfig}
\usepackage{graphicx}
\usepackage{amsmath}
\usepackage{amssymb}
\usepackage{stackengine}
\usepackage[utf8]{inputenc} % allow utf-8 input
\usepackage[T1]{fontenc}    % use 8-bit T1 fonts
\usepackage{url}            % simple URL typesetting
\usepackage{booktabs}       % professional-quality tables
\usepackage{amsfonts}       % blackboard math symbols
\usepackage{nicefrac}       % compact symbols for 1/2, etc.
\usepackage{microtype}      % microtypography
\usepackage{graphicx}
\usepackage{multirow}
\usepackage{mathtools}
\usepackage{commath}
\usepackage{relsize}
\usepackage{tabularx}
\usepackage{makecell}

\newcommand{\vpara}[1]
{\vspace{0.1in}\noindent\textbf{#1 }}
\newcommand\blfootnote[1]{%
  \begingroup
  \renewcommand\thefootnote{}\footnote{#1}%
  \addtocounter{footnote}{-1}%
  \endgroup
}

\begin{document}
\title{Deep Fundamental Matrix Estimation without Correspondences} 
% Replace with your title

\titlerunning{Deep Fundamental Matrix Estimation without Correspondences}
% Replace with a meaningful short version of your title
%
\author{{Omid Poursaeed}\textsuperscript{*} \inst{1,2} \and
{Guandao Yang}\textsuperscript{*}\inst{1} \and
{Aditya Prakash}\textsuperscript{*}\inst{3} \and
Qiuren Fang\inst{1} \and
Hanqing Jiang\inst{1} \and
Bharath Hariharan\inst{1} \and
Serge Belongie\inst{1,2}
}
%
%Please write out author names in full in the paper, i.e. full given and family names. 
%If any authors have names that can be parsed into FirstName LastName in multiple ways, please include the correct parsing, in a comment to the volume editors:
%\index{Lastnames, Firstnames}
%(Do not uncomment it, because you may introduce extra index items if you do that, we will use scripts for introducing index entries...)
\authorrunning{O. Poursaeed, G. Yang, A. Prakash, Q. Feng, and H. Jiang, B. Hariharan, S. Belongie}
% Replace with shorter version of the author list. If there are more authors than fits a line, please use A. Author et al.
%

\institute{Cornell University \and
Cornell Tech \and
Indian Institute of Technology Roorkee}
\maketitle              % typeset the header of the contribution
\blfootnote{* Indicates equal contribution}
\begin{abstract}
%\blfootnote{* Indicates equal contribution}
Estimating fundamental matrices is a classic problem in computer vision. %It describes the geometric relations that exist between two images of the same scene. 
Traditional methods rely heavily on the correctness of estimated key-point correspondences, which can be noisy and unreliable. 
As a result, it is difficult for these methods to handle image pairs with large occlusion or significantly different camera poses. In this paper, we propose novel neural network architectures to estimate fundamental matrices in an end-to-end manner without relying on point correspondences. New modules and layers are introduced in order to preserve mathematical properties of the fundamental matrix as a homogeneous rank-2 matrix with seven degrees of freedom. We analyze performance of the proposed models using various metrics on the KITTI dataset, and show that they achieve competitive performance with traditional methods without the need for extracting correspondences.

\keywords{Fundamental Matrix \and Epipolar Geometry \and Deep Learning \and Stereo.}
\end{abstract}
The Fundamental matrix (F-matrix) contains rich information relating two stereo images. %, including relative rotation, translation and camera intrinsics. 
The ability to estimate fundamental matrices is essential for many computer vision applications such as camera calibration and localization, image rectification, depth estimation and 3D reconstruction. 
The current approach to this problem is based on detecting and matching local feature points, and using the obtained correspondences to compute the fundamental matrix by solving an optimization problem about the epipolar constraints \cite{longuet1981computer,hartley2003multiple}.
The performance of such methods is highly dependent on the accuracy of the local feature matches, which are based on algorithms such as SIFT~\cite{sift}. 
However, these methods are not always reliable, especially when there is occlusion, large translation or rotation between images of the scene. 

In this paper, we propose end-to-end trainable convolutional neural networks for F-matrix estimation that do not rely on key-point correspondences. 
 The main challenge of directly regressing the entries of the F-matrix is to preserve its mathematical properties as a homogeneous rank-2 matrix with seven degrees of freedom. We propose a reconstruction module and a normalization layer (Sec.~\ref{sec:normalization}) to address this challenge. We demonstrate that by using these layers, we can accurately estimate the fundamental matrix, while a simple regression approach does not yield good results. Our detailed network architectures are presented in Sec.~\ref{sec:architecture}.
 Empirical experiments are performed on the KITTI dataset~\cite{geiger2013vision} in Sec.~\ref{sec:experiments}. The results indicate that we can achieve competitive results with traditional methods without relying on correspondences.

% In brief, our contributions are as follows:
% \begin{itemize}
% \item We used deep learning to capture important semantic features compared to the traditional methods.
% \item Our model leveraged an F-matrix reconstruction module that could preserve the mathematical property of F-matrix.
% \item We proposed a new evaluation metric that is based on epipolar geometric, and does not depend on key-point correspondences.
% \end{itemize}
% The rest of the paper is organized as follows. Section 2 reviews the related work. Section 3 explains our proposed approach and our CNN model and architecture. Section 4 discusses in detail the experimental setup, parameter settings and our results.

% \vspace{-1em}
\section{Background and Related Work}
\label{sec:related-works}
% \vspace{-0.5em}

% Traditional Methods:
% 1. key-point correspondence based methods
% 2. minimizing other loss
% 3. Difference: they are not resilient to noise, relied on the key-point pairs, no deep learning
\subsection{Fundamental Matrix and Epipolar Geometry}

When two cameras view the same 3D scene from different viewpoints, geometric relations among the 3D points and their projections onto the 2D plane lead to constraints on the image points. This intrinsic projective geometry is referred to as the epipolar geometry, and is encapsulated by the fundamental matrix $\mathbf{F}$. %, and is 
%independent of scene structure. 
This matrix only depends on the cameras’ internal parameters and their relative pose, and can be computed as:
\begin{equation}\label{eq:f-def}
    \mathbf{F} = \mathbf{K_2}^{-T} [\mathbf{t}]_{\times} \mathbf{R}  \mathbf{K_1}^{-1}
\end{equation}
where $\mathbf{K_1}$ and $\mathbf{K_2}$ represent camera intrinsics, and $\mathbf{R}$ and $[\mathbf{t}]_{\times}$ are the relative camera rotation and translation respectively \cite{hartley2003multiple}. More specifically:
\begin{equation}
 \begin{aligned}
    \mathbf{K}_i = \begin{bmatrix}
    f_i^{-1} & 0 & c_x \\
    0 & f_i^{-1} & c_y \\
    0 & 0 & 1
    \end{bmatrix}
   \end{aligned}
\end{equation}
\begin{equation}
\begin{aligned}
\mathbf{t}_{ \times} = \begin{bmatrix}
    0 & -t_z & t_y \\
    t_z & 0 & -t_x \\
    -t_y & t_x & 0
    \end{bmatrix}
\end{aligned}
\end{equation}
\begin{equation}
\mathbf{R} = \mathbf{R_x}(r_x) \mathbf{R_y}(r_y) \mathbf{R_z}(r_z)
\end{equation}
in which $(c_x, c_y)^T$ is the principal point of the camera, $f_i$ is the focal length of camera $i=1, 2$, and $t_x$, $t_y$ and $t_z$ are the relative displacements along the $x$, $y$ and $z$ axes respectively. $\mathbf{R}$ is the rotation matrix which can be decomposed into rotations along $x$, $y$ and $z$ axes. We assume that the principal point is in the middle of the image plane.

While the fundamental matrix is independent of the scene structure, it can be computed
from correspondences of projected scene points alone, without requiring knowledge of the cameras’ internal parameters or relative pose.
If $p$ and $q$ are matching points in two stereo images, the fundamental matrix $\mathbf{F}$ satisfies the equation:
\begin{equation}\label{eq:f-matrix}
q^T \mathbf{F} p = 0
\end{equation}
Writing $p = (x, y, 1)^T$ and $q= (x', y', 1)^T$ and $\mathbf{F}=[f_{ij}]$, equation \ref{eq:f-matrix} can be written as:
\begin{equation}\label{eq:linear}
x'x f_{11}+ x'y f_{12}+ x' f_{13}+ y'x f_{21}+ y'y f_{22}+y' f_{23}+xf_{31}+y f_{32}+f_{33} = 0.
\end{equation}
Let $\mathbf{f}$ represent the 9-vector made up of the entries of $\mathbf{F}$. Then equation \ref{eq:linear} can be written as: 
\begin{equation}
(x'x, x'y, x', y'x, y'y, y', x, y, 1) \mathbf{f} = 0
\end{equation}
A set of linear equations can be obtained from $n$ point correspondences:
\begin{equation}\label{eq:matrix}
\mathbf{A} \mathbf{f} =  
\begin{bmatrix}
x_1'x_1 & x_1'y_1 & x_1' & y_1'x_1 & y_1'y_1 & y_1' & x_1 & y_1 & 1 \\
\vdots & \vdots & \vdots & \vdots & \vdots & \vdots & \vdots & \vdots & \vdots \\
x_n'x_n & x_n'y_n & x_n' & y_n'x_n & y_n'y_n & y_n' & x_n & y_n & 1
\end{bmatrix}
\mathbf{f} = 0
\end{equation}

Various methods have been proposed for estimating fundamental matrices based on equation \ref{eq:matrix}. 
The simplest method is the eight-point algorithm which was proposed by Longuet-Higgins~\cite{longuet1981computer}. 
Using (at least) 8 point correspondences, it computes a (least-squares) solution to equation \ref{eq:matrix}. It enforces the rank-2 constraint using Singular Value Decomposition (SVD), and finds a matrix with the minimum Frobenius distance to the computed (rank-3) solution.  
%of a set of linear equations based on equation \ref{eq:f-matrix}.
Hartley~\cite{hartley1997defense} proposed a normalized version of the eight-point algorithm which achieves improved results and better stability. The algorithm involves translation and scaling of the points in the image before formulating the linear equation \ref{eq:matrix}.

The Algebraic Minimization algorithm uses a different procedure for enforcing the rank-2 constraint. It tries to minimize the algebraic error $\mathbf{A} \norm{\mathbf{f}}$ subject to $\norm{\mathbf{f}}=1$. It uses the fact that we can write the singular fundamental matrix as $\mathbf{F}=\mathbf{M}[e]_\times$ where $\mathbf{M}$ is a non-singular matrix and $[e]_\times$ is a skew-symmetric matrix with $e$ corresponding to the epipole in the first image. This equation can be written as $\mathbf{f}=E\mathbf{m}$, where $\mathbf{f}$ and $\mathbf{m}$ are vectors comprised of entries of $\mathbf{F}$ and $\mathbf{M}$, and $E$ is a $9\times 9$ matrix comprised of elements of $[e]_\times$. Then the minimization problem becomes: 
\begin{equation}
\textrm{minimize } \norm{\mathbf{A}E\mathbf{m}} \textrm{ subject to } \norm{E\mathbf{m}}=1
\end{equation}
To solve this optimization problem, we can start from an initial estimate of $\mathbf{F}$ and set $e$ as the generator of the right null space of $\mathbf{F}$. Then we can iteratively update $e$ and $\mathbf{F}$ to minimize the algebraic error. More details are given in \cite{hartley2003multiple}. 

The Gold Standard geometric algorithm assumes that the noise in image point measurements obeys a Gaussian distribution. It tries to find the Maximum Likelihood estimate of the fundamental matrix which minimizes the geometric distance
\begin{equation}
{\sum_i}{d(p_i, \hat{p}_i)^2 +  d(q_i, \hat{q}_i)^2} 
\end{equation}
in which $p_i$ and $q_i$ are true correspondences satisfying equation \ref{eq:f-matrix}, and $\hat{p}_i$ and $\hat{q}_i$ are the estimated correspondences.  

Another algorithm uses RANSAC~\cite{fischler1981random} to compute the fundamental matrix. It computes interest points in each image, and finds correspondences based on proximity and similarity of their intensity neighborhood. In each iteration, it randomly samples 7 correspondences and computes the F-matrix based on them. It then calculates the re-projection error for each correspondence, and counts the number of inliers for which the error is less than a specified threshold. After sufficient number of iterations, it chooses the F-matrix with the largest number of inliers. A generalization of RANSAC is MLESAC~\cite{torr2000mlesac}, which adopts the same sampling strategy as RANSAC to generate putative solutions, but chooses the solution that maximizes the likelihood rather than just the number of inliers. MAPSAC ~\cite{torr2002bayesian} (Maximum A Posteriori SAmple Consensus) improves MLESAC by being more robust against noise and outliers including Bayesian probabilities in minimization.
A global search genetic algorithm combined with a local search hill climbing algorithm is proposed in~\cite{zhang2016fundamental} to optimize MAPSAC algorithm for estimating fundamental matrices. 
%~\cite{xiao2018efficient} introduces a robust estimation method called inlier set sample optimization for fundamental matrix estimation of wide baseline images. 
\cite{yan2014fundamental} proposes an algorithm to cope with the problem of fundamental matrix estimation for binocular vision system used in wild field. It first acquires the edge points using Canny edge detector, and then gets the pre-matched points by the GMM-based point set registration algorithm. It then computes the fundamental matrix using the RANSAC algorithm.
~\cite{fathy2007essential} proposes to use adaptive penalty methods for valid estimation of Essential matrices as a product of translation and rotation matrices.  
A new technique for calculating the fundamental matrix combined with feature lines is introduced in~\cite{zhou2015method}. The interested reader is referred to~\cite{armangue2003overall} for a survey of various methods for estimating the F-matrix.

%A global search genetic algorithm combining with a local search hill climbing algorithm is proposed in~\cite{zhang2016fundamental} to optimize MAPSAC algorithm for estimating fundamental matrices.

% was used to find inliers and get a more robust estimation. These methods minimize the re-projection error, Sampson distance and other loss functions. Key-point correspondences are mostly computed from hand-crafted feature extractors like SIFT~\cite{sift}. These models tend to fail in the case where viewpoints are drastically different or where higher-level features are needed to make correct correspondence decisions. {\bf UCN}

\subsection{Deep Learning for Multi-view Geometry} 
Deep neural networks have achieved state-of-the-art performance on tasks such as image recognition~\cite{krizhevsky2012imagenet,he2016deep,szegedy2015going,simonyan2014very}, semantic segmentation~\cite{long2015fully,chen2018deeplab,yu2015multi,zhao2017pyramid}, object detection~\cite{girshick2014rich,ren2015faster,redmon2016you}, scene understanding~\cite{kendall2015bayesian,zhou2016places,poursaeed2017vision} and generative modeling~\cite{goodfellow2014generative,radford2015unsupervised,huang2017stacked,zhang2017stackgan,poursaeed2017generative} in the last few years. Recently, there has been a surge of interest in using deep learning for classic geometric problems in Computer Vision.
A method for estimating relative camera pose using convolutional neural networks is presented in~\cite{melekhov2017relative}.  
It uses a simple convolutional network with spatial pyramid pooling and fully connected layers to compute the relative rotation and translation of the camera. 
An approach for camera re-localization is presented in~\cite{laskar2017camera} which localizes a given query image by using a convolutional neural network for first retrieving similar database images and then predicting the relative pose between the query and the database images with known poses. The camera location for the query image is obtained via triangulation from two relative translation estimates using a RANSAC-based approach.
~\cite{workman2015deepfocal} uses a deep convolutional neural network to directly estimate the focal length of the camera using only raw pixel intensities as input features.
~\cite{brachmann2017dsac} proposes two strategies for differentiating the RANSAC algorithm: using a soft argmax operator, and probabilistic selection. 
\cite{garon2017deep} leverages deep neural networks for 6-DOF tracking of rigid objects. 

~\cite{deephomography} presents a deep convolutional neural network for estimating the relative homography between a pair of images. A more complicated algorithm is proposed in~\cite{erlik2017homography} which contains a hierarchy of twin convolutional regression networks to estimate the homography between a pair of images. ~\cite{detone2017toward} introduces two deep convolutional neural networks, MagicPoint and MagicWarp. MagicPoint extracts salient 2D points from a single image. MagicWarp operates on pairs of point images (outputs of MagicPoint), and estimates the homography that relates the inputs. 
~\cite{nguyen2018unsupervised} proposes an unsupervised learning algorithm that trains a deep convolutional neural network to estimate planar homographies.
A self-supervised framework for training interest point detectors and descriptors is presented in~\cite{detone2017superpoint}.  
A convolutional neural network architecture for geometric matching is proposed in~\cite{rocco2017convolutional}. It uses feature extraction networks with shared weights and a matching network which matches the descriptors. The output of the matching network is passed through a regression network which outputs the parameters of the geometric transformation. 
~\cite{ji2017surfacenet} presents a model which takes a set of images and their corresponding camera parameters as input and directly infers the 3D model.

\section{Network Architecture}
\label{sec:methods}
\label{sec:architecture}
% 1. motivation ConvNets could automatically learn feature that's useful
%   i)  it could grab higher order feature
%   ii) higher order feature should be able to handle problem of large occlusion, translation, or rotation
% 2. Detailed introduction of the architecture, ConvNets with downsampling, Cite HomographyNet

We leverage deep neural networks for estimating the fundamental matrix directly from a pair of stereo images. 
Each network consists of a feature extractor to obtain features from the images and a regression network to compute the entries of the F-matrix from the features. %In the following sections, we describe each part separately.
\subsection{Feature Extraction}
\begin{figure*}[t!]
% \vspace{-6mm}
\centering
\includegraphics[width=0.9\linewidth]{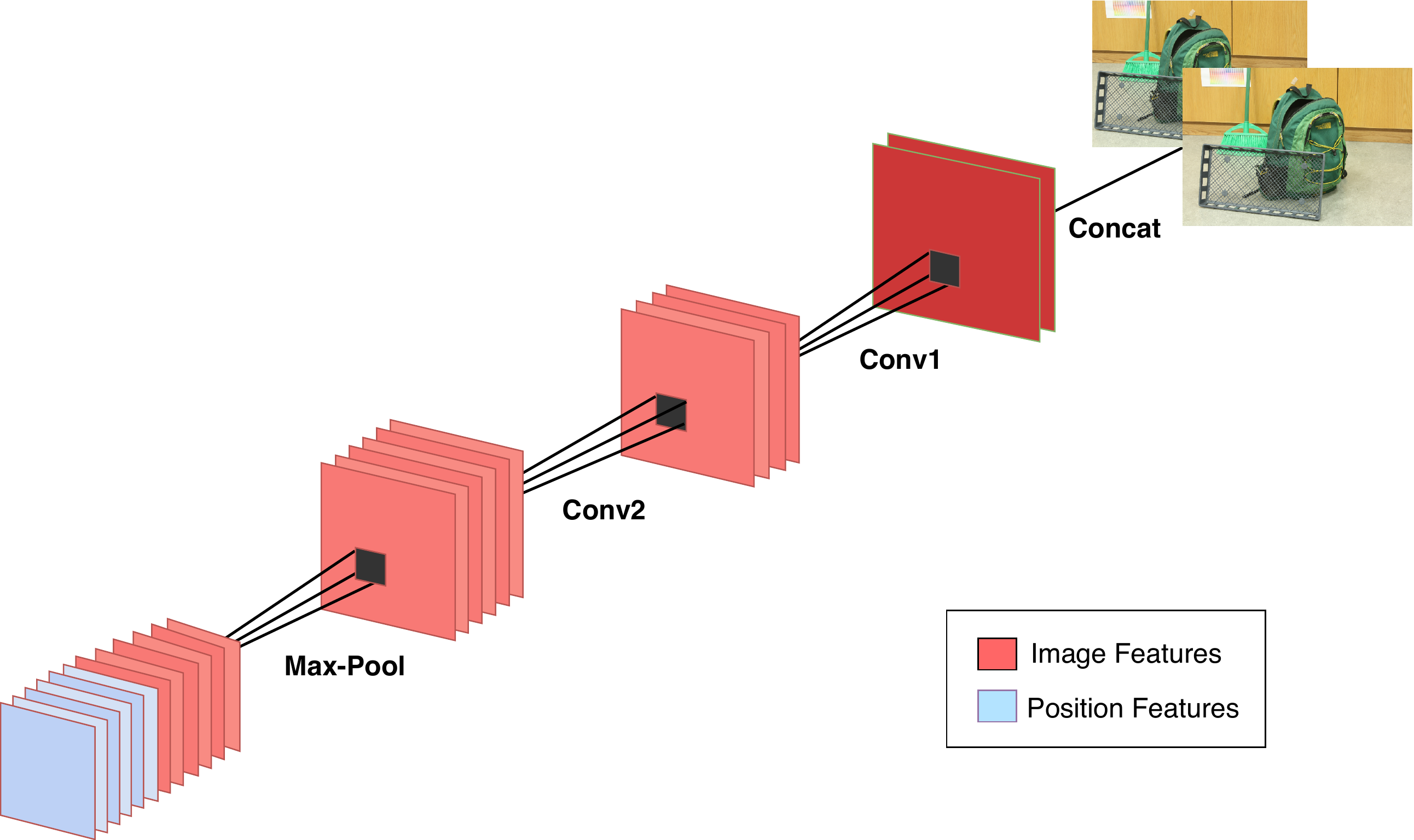}
\centering
%\vspace{-6mm}
\caption{Single-Stream Architecture. Stereo images are concatenated and passed to a convolutional neural network. Position features can be used to indicate where the final activations come from with respect to the full-size image. 
}
\label{fig:single-network}
%  \vspace{-1mm}
\end{figure*}

We consider two different architectures for feature extraction. 
In the first architecture, we concatenate the images across the channel dimension, and pass the result to a {neural network} to extract features.  
%since such networks are good at extracting higher level features. 
Figure~\ref{fig:single-network} illustrates the network structure. {We use two convolutional layers, each followed by ReLU and Batch Normalization~\cite{ioffe2015batch}. We use 128 filters of size $3\times 3$ in the first convolutional layer and 128 filters of size $1\times 1$ in the second layer. We limit the number of pooling layers to one in order not to lose the spatial structure in the images.} %The extracted features can then be processed by the regression network. %The resulting features are relatively robust to occlusion, large translations and rotations between the stereo images.  

\begin{figure}
% \vspace{-6mm}
\centering
\includegraphics[width=\linewidth]{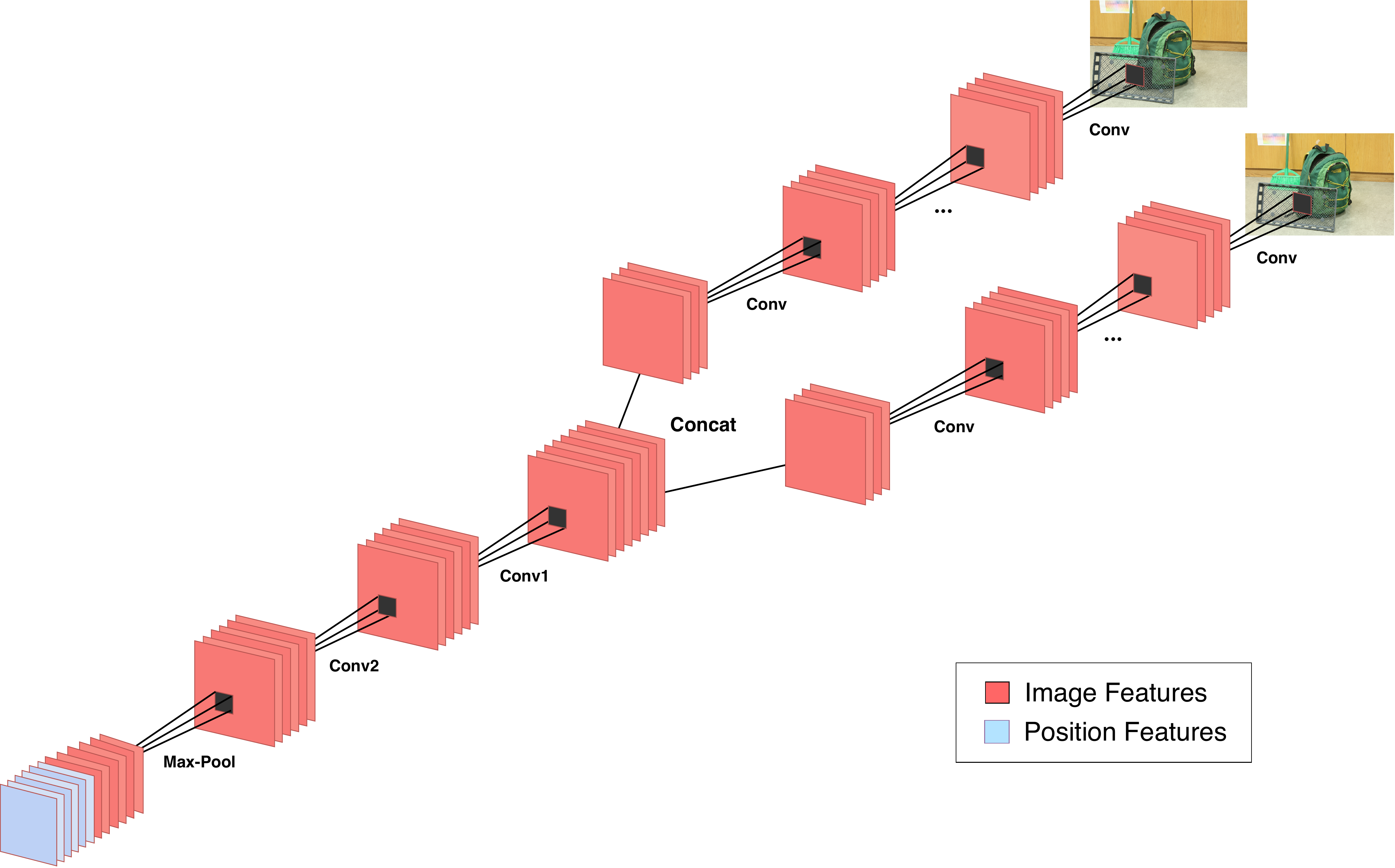}
\centering
%\vspace{-6mm}
\caption{Siamese Architecture. Images are first passed to two streams with shared weights. The resulting features are concatenated and passed to the single-stream network as in figure \ref{fig:single-network}. Position features can be used with respect to the concatenated features. }
\label{fig:siamese-network}
%  \vspace{-1mm}
\end{figure}
% We use five convolution groups, down-sampling the input images from $256\times 256$ to $16\times16$. The number of filters within each group is $64$, $64$, $128$, $128$, and $128$ respectively. Connecting groups are maximum pooling layers with pool-size $2\times2$.

\vpara{Location Aware Pooling.}
\label{sec:location-aware-max-pooling}
% State the problem:
%   i)   During MaxPooling's downsampling, locations of the activated features are lost.
%   ii)  The convolution layer is also location invariant, making it hard to know where a certain activate is
%   iii) The location of the activation, however, should be the most useful feature to compute the F-matrix.
% State the solution:
%   i)   We index the image with (1...256*256)/(256*256) so that it's within 0 and 1
%   ii)  We record the Max Pooling's activation location along the way
%   iii) Make some formular
%   iv)  We have detailed experiment data to compare the effectiveness of such features.
As discussed in Sec. \ref{sec:related-works}, the F-matrix is highly dependent on the relative location of corresponding points in the images. However, down-sampling layers such as Max Pooling discard the location information. In order to retain this information,  
%Another challenge comes from the fact that the features extracted by the network are agnostic to locations, especially when these features are down-sampled in the max-pooling layer. The locations of these activation, however, are useful for computing the F-matrix. 
we keep all the indices of where the activations come from in the max-pooling layers. At the end of the network, we append the position of final features with respect to the full-size image. Each location is indexed with an integer in $[1, h\times w \times c]$ normalized to be within the range $[0,1]$, in which {$h$, $w$ and $c$ are the height, width and channel dimensions of the image respectively.} In this way, each feature has a position index indicating from where it comes from. This helps the network to retain the location information and to provide more accurate estimates of the F-matrix. %As a result, the output feature vector will double its depth from $16\times16\times128$ to $16\times16\times256$.

The second architecture is shown in figure \ref{fig:siamese-network}. We first process each of the input images in a separate stream using an architecture similar to the Universal Correspondence Network (UCN)~\cite{choy2016universal}. 
{Unlike the UCN architecture, we do not use Spatial Transformers~\cite{jaderberg2015spatial} in these streams since they can remove part of the information needed for estimating relative camera rotation and translation.} The resulting features from these streams are then concatenated, and passed to a single-stream network similar to {figure \ref{fig:single-network}}. 
We can use position features in the single-stream network as discussed previously. These features capture the position of final features the with respect to the concatenated features at the end of the two streams. We refer to this architecture as `Siamese'. As we show in Sec. \ref{sec:experiments}, this network outperforms the Single-Stream one. We also consider using only the UCN without the single-stream network. The results, however, are not competitive with the Siamese architecture. 
%processing images  can help us . 
%The extracted features from each stream are concatenated, and passed to the regression network.   

\subsection{Regression}

A simple approach 
for computing the fundamental matrix from the features is to pass them to fully-connected layers, and directly regress the nine entries of the F-Matrix. We can then normalize the result to achieve scale-invariance. This approach is shown in figure \ref{fig:regression} (left). 
The main issue with this approach is that the predicted matrix might not satisfy all the mathematical properties required for a fundamental matrix as a rank-2 matrix with seven degrees of freedom. %These properties cannot be easily enforced by a model that directly regresses the 9 matrix entries. 
In order to address this issue, we introduce Reconstruction and Normalization layers in the following. 

%The final image features will be used in two ways: 1) put into a MLP to produce 9 parameters, and these 9 parameters will be put into a normalization layer and output a normalized matrix as the prediction; and 2) put into a MLP to predict 8 parameters, and these 8 parameters will be used to reconstruct a F-matrix using the reconstruction layer and the normalization layer. The reconstruction layer is presented in Sec.~\ref{sec:reconstruction-module}, and the normalization layer is presented in Sec.~\ref{sec:normalization}.
\subsubsection{F-matrix Reconstruction Layer. }
\label{sec:reconstruction-module}

% 1. State the problem
%    i) F-matrix has many mathematical properties, such as 
%        a) rank 2 matrix
%        b) F transpose is the reverse matrix
%        c) 2-degree of freedom
%   Such properties is hard to be directly capture by the network through direct regression to the ground truth
% 2. Solution:
%   i) We know that the fundamental matrix is constructed through the following formular:
%   F = K2^T*R*[t]x*K1^T
%   ii) We would like to directly regress the 8 parameters : f1, f2, rx, ry, rz, tx, ty, tz;
%   iii) then we use these parameters to reconstruct the F-matrix
%   iv)  the final f-matrix will be pull toward the ground truth through L2 loss

\begin{figure*}[t!]
% \vspace{-6mm}
%\begin{center}
\centering
% \includegraphics[width=\linewidth]{figures/model.png}
%\begin{subfigure}{0.93\textwidth}
\includegraphics[width=0.425\linewidth]{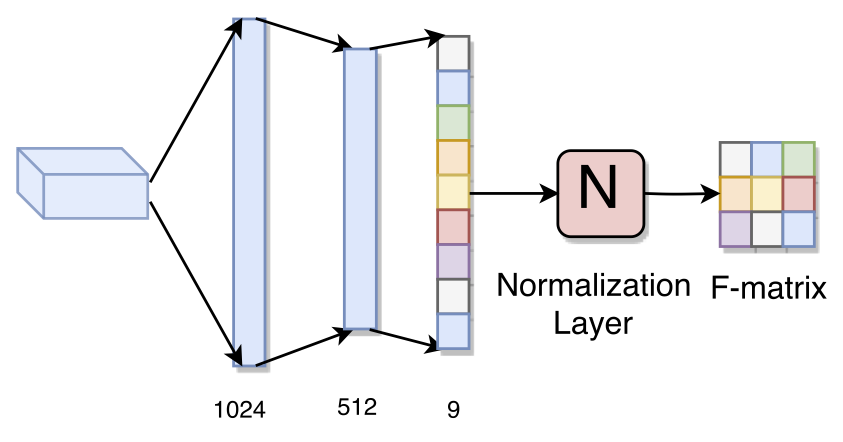}
\label{fig:direct-regression}
%\end{subfigure}
%\begin{subfigure}
\includegraphics[width=0.525\linewidth]{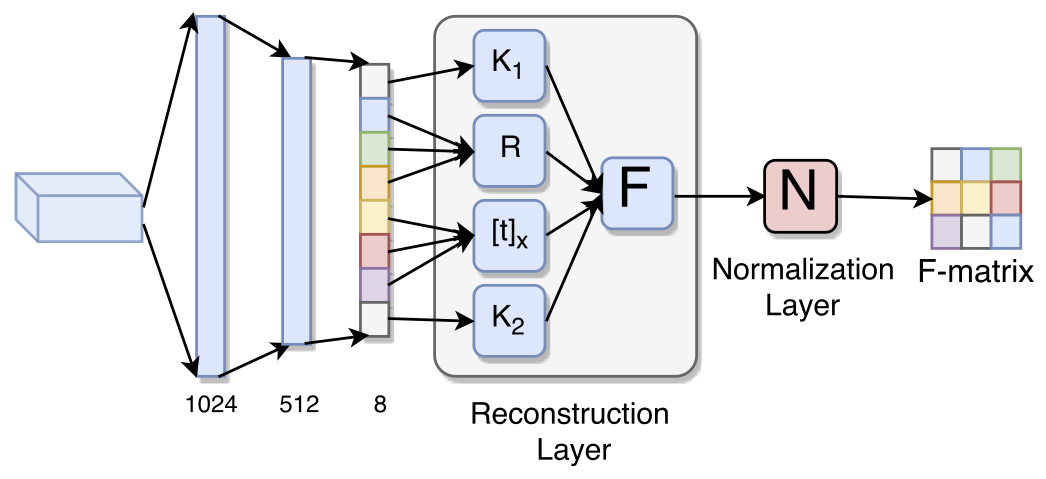}
\label{fig:reconstruction}
%\end{subfigure}
%\end{center}
% \vspace{-6mm}
\caption{
Different regression methods for predicting F-matrix entries from the features. The architecture to directly regress the entries of the F-matrix is shown on the left. The network with the reconstruction and normalization layers is shown on the right, and is able to estimate homogeneous F-matrices with rank two and seven degrees of freedom. 
}
\label{fig:regression}
\vspace{-4mm}
\end{figure*}

We consider equation \ref{eq:f-def} to reconstruct the fundamental matrix:
\begin{equation}
    \mathbf{\hat{F}} = \mathbf{K_2}^{-T} [\mathbf{t}]_{\times} \mathbf{R}  \mathbf{K_1}^{-1}
\end{equation}
we need to determine eight parameters $(f_1, f_2, t_x, t_y, t_z, r_x, r_y, r_z)$ as shown in equations (\textcolor{red}{2--4}). 
Note that the predicted $\mathbf{\hat{F}}$ is differentiable with respect to these parameters. Hence, we can construct a layer that takes these parameters as input, and outputs a fundamental matrix $\mathbf{\hat{F}}$. This approach guarantees that the reconstructed matrix has rank two. Figure \ref{fig:regression} (right) illustrates the Reconstruction layer.

\subsubsection{Normalization Layer. }
\label{sec:normalization}

% State the problem:
% * Since the F-matrix has invariant from the scale, so in order to compare two F-matrix, we need to normalize it first. We compared three ways to normalize the F-matrix:
% 1. divided by the last entry : most common way, but usually result in some exceptionally large values in other entries.
% 2. divided by the L_2 norm (Fobinius norm) : it will live in a higher-dimensional sphere, make the prediction space smaller
% 3. divided by the maximum absolute value -> restricted all number to be within [-1,1]
% We provide empirical details to compare among these different normalization methods.

Considering that the F-matrix is scale-invariant, we also use a Normalization layer to remove another degree of freedom for scaling. In this way, the estimated F-matrix will have seven degrees of freedom and rank two as desired. 
%In order to make the predicted fundamental matrix comparable with the ground-truth, we have to eliminate the scale factor, 
The common practice for normalization is to divide the F-matrix by its last entry. We call this method \textbf{ETR-Norm}. However, since the last entry of the F-matrix could be close to zero, this can result in large entries, and training can become unstable. Therefore, we propose two alternative normalization methods. 

\vpara{FBN-Norm:} We divide all entries of the F-matrix by its Frobenius norm, so that all the matrices live on a 9-sphere of unit norm. Let $\|\mathbf{F}\|_F$ denote the Frobenius norm of matrix $\mathbf{F}$. Then the normalized fundamental matrix is: 
\begin{equation}
    \mathcal{N}_{FBN}(\mathbf{F}) = \|\mathbf{F}\|_{F}^{-1}\mathbf{F}
\end{equation}

\vpara{ABS-Norm:} We divide all entries of the F-matrix by its maximum absolute value, so that all entries are restricted within $[-1,1]$ range:
\begin{equation}
    \mathcal{N}_{ABS}(\mathbf{F}) = (\max_{i,j}|\mathbf{F}_{i,j}|)^{-1}\mathbf{F}
\end{equation}

During training, the normalized F-matrices are compared with the ground-truth using both $L_1$ and $L_2$ losses. We provide empirical results to study how each of these normalization methods influences performance and stability of training in Sec.~\ref{sec:experiments}. 

\subsubsection{Epipolar Parametrization}
\label{sec:epipolar}
Given that the F-matrix has a rank of two, an alternative parametrization is specifying the first two columns $\mathbf{f}_1$ and $\mathbf{f}_2$ and the coefficients $\alpha$ and $\beta$ such that $\mathbf{f}_3 = \alpha \mathbf{f}_1 + \beta \mathbf{f}_2 $. Normalization layer can still be used to achieve scale-invariance. The coordinates of the epipole occur explicitly in this parametrization: $(\alpha, \beta, 1)^T$ is the right epipole for the F-matrix~\cite{hartley2003multiple}. The corresponding regression architecture is similar to figure~\ref{fig:regression}, but we interpret the final eight values differently: the first six elements represent the first two columns and the last two represent the coefficient for combining the columns. 
%As we will show in Sec.~\ref{sec:results} this parametrization works particularly well for normalizing with respect to the last entry (ETR-Norm).
The main disadvantage of this method is that it does not work when the first two columns of $\mathbf{F}$ are linearly dependent. In this case, it is not possible to write the third column in terms of the first two columns. 

% \subsection{Training}
% \label{sec:training}
% 
% % 1. Training is composed of both L_2 and L_1 losses
% % Motivation:
% % * normally L2 loss is what we use
% % * L2 loss make great penalty in the outliners, but easily get stuck into the mean of the distribution, yet F-matrix might not be simply a unimodal distribution.
% % * but adding L_1 loss make the gradient flow stronger and improve the regression precision.
% 
% We use both $L_1$ and $L_2$ loss between the predicted F-matrix and the ground truth F-matrix to train the model. Note that the ground truth F-matrix will be normalized in the same way as the ones output by the model. One advantage of using $L_1$ loss is that it helps to regularize the model, and it improve performance when used in combination with the $L_2$ loss, which will penalize strong out-liners. The other advantage of the $L_1$ loss is that it keeps the gradient flow from diminishing. Our model used Adam optimizer (CITE), and we clipped the gradient to have $L_2$ norm of $1$ to stabalize the training.

\section{Experiments}
\label{sec:experiments}

To evaluate whether our models can successfully learn F-matrices, we train models with various configurations and compare their performance based on the metrics defined in Sec.~\ref{sec:evaluation}. 
The baseline model (\textbf{Base}) uses neither position features nor the reconstruction module. The \textbf{POS} model utilizes the position features on top of the \textbf{Base} model.  
Epipolar parametrization (Sec.~\ref{sec:epipolar}) is used for the \textbf{EPI} model. \textbf{EPI+POS} uses the position features with epipolar parametrization. 
The \textbf{REC} model is the same as \textbf{Base} but uses the reconstruction module. Finally, the \textbf{REC+POS} model uses both the position features and the reconstruction module.

{We use the KITTI dataset for training our models.} The dataset has been recorded from a moving platform while driving in and around Karlsruhe, Germany. We use $2000$ images from the raw stereo data in the `City' category, and split them into 1600 train, $200$ validation and $200$ test images. Ground truth F-matrices are obtained using the ground-truth camera parameters. The same normalization methods are used for both the estimated and the ground truth F-matrices. %These matrices are compared using the $L_2$ loss.
The feature extractor and the regression network are trained jointly in an end-to-end manner.  

%Results are compared in Table~\ref{table:results}.

%\subsection{Dataset}
%\label{sec:dataset}
% Briefly discribe how we generate dataset
% 1. single scene
% 2. PovRay render
% 3. Use OpenCV to generate the groundtruth
% 4. keep the key-points

%In order to obtain ground-truth F-matrices for training, we develop a synthetic dataset based on POV-Ray~\cite{povray}. We use a simple scene to render roughly 600 different images, all of which could see the center of the scene. Then we pair up all the images and compute the ground truth Fundamental matrices and key-point correspondences between two images using OpenCV~\cite{opencv}. Note that only a small part of key-point correspondences are used to compute the ground-truth using either seven-pointscontains wide-baseline stereo images  or eight-points algorithm, and the remaining points are held out for evaluation.
%We consider the KITTI datasets for training our models.
%:  KITTI~\cite{kitti} and ALOI~\cite{geusebroek2005amsterdam}. 
% The KITTI dataset has been recorded from a moving platform while driving in and around Karlsruhe, Germany. 

% The Amsterdam Library of Object Images (ALOI) is a collection of 1,000 objects recorded under various imaging circumstances. It contains 750 wide-baseline stereo images. By turning the object 15 degrees to view the second camera, it obtains a center image, a right image and a left image. The combination of left-center and center-right images yields two pairs of 15 degrees-baseline stereo, whereas the combination of the left-right pair yields a 30 degrees-baseline stereo image. 

\subsection{Evaluation Metrics}
\label{sec:evaluation}
% Briefly explain how we like to evaluate the matric
% 1. epipolar constriant absolute value
% 2. epipolar constraint squred
% 3. Sampson Distance
% 4. Symmetrical Epipolar Distances

We use the following metrics to measure how well the F-matrix satisfies the epipolar constraint (equation \ref{eq:f-matrix}) according to the held out correspondences: 
%To evaluate both the ground truth and the predicted F-matrices, 

\vpara{EPI-ABS (Epipolar Constraint with Absolute Value):} 
\begin{equation}\label{eq:epi-abs}
    \mathcal{M}_{EPI-ABS}(\mathbf{F}, {p}, {q}) = \sum_{i} |{q}_i^T\mathbf{F}{p}_i|
\end{equation}

\vpara{EPI-SQR (Epipolar Constraint with Squared Value):}
\begin{equation}\label{eq:epi-sq}
\mathcal{M}_{EPI-SQR}(\mathbf{F}, {p}, {q}) = \sum_{i} ({q}_i^T\mathbf{F}{p}_i)^2    
\end{equation}   

% \textbf{SSD (Sampson Distance):}
% \begin{equation}
% \mathcal{M}_{SSD}(\mathbf{F}, \mathbf{p}, \mathbf{q}) = \sum_{i} \frac{
% (\mathbf{p}_i^T\mathbf{F}\mathbf{q}_i)^2
% }{
% (\mathbf{F}\mathbf{p}_i)^2_1 + 
% (\mathbf{F}\mathbf{p}_i)^2_2 +
% (\mathbf{F}\mathbf{q}_i)^2_1 +
% (\mathbf{F}\mathbf{q}_i)^2_2
% }
% \end{equation}

% \textbf{SED (Symmetrical Epipolar Distance):}
% \begin{equation}
% \mathcal{M}_{SED}(\mathbf{F}, \mathbf{p}, \mathbf{q}) = (\mathbf{p}_i^T\mathbf{F}\mathbf{q}_i)^2
% \Big(
%     \frac{1}{(\mathbf{F}\mathbf{p}_i)^2_1 + (\mathbf{F}\mathbf{p}_i)^2_2} + 
%     \frac{1}{(\mathbf{F}\mathbf{q}_i)^2_1 +(\mathbf{F}\mathbf{q}_i)^2_2}
% \Big)
% \end{equation}

The first metric is equivalent to the Algebraic Distance mentioned in~\cite{fathy2011fundamental}. We evaluate the metrics based on high-confidence key-point correspondences: we select the key-points for which the Symmetric Epipolar Distance based on the ground-truth F-matrix is less than 2 \cite{hartley2003multiple}. This ensures that the point is no more than one pixel away from the corresponding epipolar line. %Other potential metrics are Symmetrical Epipolar Distance (SED) and Sampson Distance (SSD). However, the denominator terms in these metrics can become very small for some stereo images which results in large errors for both the estimated and the ground truth F-matrices. Therefore, we focus on the metrics in equations \ref{eq:epi-abs} and \ref{eq:epi-sq} which are more stable. %We evaluate the  

\begin{table}[t!]
\centering
 \resizebox{\textwidth}{!}{
\begin{tabular}{|l|l|l|l|l|l|l|l|l|l|l l|}
\Xhline{3\arrayrulewidth}
& \multicolumn{3}{c|}{Siamese Network} & \multicolumn{3}{c|}{Single-stream Network} \\
\Xhline{3\arrayrulewidth}
Normalization & Models & {EPI-ABS} & {EPI-SQR} & Models & {EPI-ABS} & EPI-SQR  \\ 
\Xhline{3\arrayrulewidth}
{ETR-Norm} & Base & 3.77 & 27.16 & Base & 4.43 & 34.34 \\
 & POS  & 4.05 & 21.90 & POS & 2.47 & 9.79 \\ 
 & EPI & 0.52 & 0.28 & EPI & 1.00 & 0.99 \\
 & EPI + POS & 0.88 & 1.02 & EPI + POS & 1.00 & 1.00 \\
 & REC & 0.56 & 0.45 & REC & 0.99 & 0.99 \\
 & REC + POS  & 0.97 & 0.98 & REC + POS & 1.00 & 0.99 \\ %\cline{2-5} \cline{7-10} 
 \hline
 % & 7-point & 1.09 & 25.5 & 7-point & 1.09 & 25.5 \\ 
   & 8-point & 1.91 & 152.83 & 8-point & 1.91 & 152.83 \\ 
    & LeMedS & 1.09 & 25.50 & LeMedS & 1.09 & 25.50 \\ 
 & RANSAC & 0.60 & 3.85 & RANSAC & 0.60 & 3.85 \\ 
 \hline
 & Ground-truth & 0.05 & 0.004 & Ground-truth & 0.05 & 0.004 \\
 \Xhline{3\arrayrulewidth}
{FBN-Norm}     & Base & 1.44 & 2.58 & Base & 2.45 & 9.99 \\
  & POS & 1.97 & 5.66 & POS & 2.78 & 8.55 \\
   & EPI & 0.07  & 0.01 & EPI & 0.91 & 0.91 \\
    & EPI + POS & 0.06 & 0.005 & EPI + POS & 0.67 & 0.58 \\
 & REC & 0.92 & 1.11  & REC & 0.78 & 1.24   \\
 & REC + POS & 0.43 & 0.44  & REC + POS & 0.87 & 0.81 \\  %\cline{2-5} \cline{7-10} 
  \hline
  %& 7-point & 0.53 & 2.93 & 7-point & 0.53 & 2.93 \\ 
  & 8-point & 1.06 & 11.7 & 8-point & 1.06 & 11.7 \\ 
   & LeMedS & 0.39 & 0.68 & LeMedS & 0.39 & 0.68 \\ 
 & RANSAC & 0.27 & 0.21 & RANSAC & 0.27 & 0.21  \\ 
 \hline
 & Ground-truth & 0.05 & 0.004 & Ground-truth & 0.05 & 0.004 \\
 \Xhline{3\arrayrulewidth}
{ABS-Norm} & Base & 4.76 & 30.63 & Base & 3.55 & 18.04 \\
 & POS & 3.74 & 22.59 & POS & 2.87 & 10.4 \\
  & EPI & 0.18 & 0.06 & EPI & 0.92 & 1.94 \\
  & EPI + POS & 0.12 & 0.03 & EPI + POS & 0.82 & 0.77  \\
 & REC & 0.22 & 0.06 & REC & 0.77 & 0.99  \\
 & REC + POS & 0.28 & 0.10 & REC + POS & 0.87  & 0.81   \\ \hline %\cline{2-5} \cline{7-10} 
  %& 7-point & 0.55 & 1.36 & 7-point & 0.55 & 1.36 \\ 
    & 8-point & 1.17 & 15.4 & 8-point & 1.17 & 15.4 \\
    & LeMedS & 0.72 & 3.88 & LeMedS & 0.72 & 3.88 \\ 
 & RANSAC & 0.33 & 0.39 & RANSAC & 0.33 & 0.39 \\
 \hline
  & Ground-truth & 0.05 & 0.004 & Ground-truth & 0.05 & 0.004 \\
 \Xhline{3\arrayrulewidth}
\end{tabular}%
}
\vspace{1em}
\caption{Results for Siamese and Single-stream networks on the KITTI dataset. Traditional methods such as 8-point, LeMedS and RANSAC are compared with different variants of our proposed model. 
Various normalization methods and evaluation metrics are considered. }
\vspace{-1em}
\label{table:results}
\end{table}

\section{Results and Discussion}
\label{sec:results}

{Results are shown in Table~\ref{table:results}}. %In terms of the normalization methods, we can see that normalizing by the Frobinius Norm results in the least error when computed the metrics in the labels the network is training on (i.e., the F-matrix computed from OpenCV). However, due to the fact that the F-matrix is scale-invariant, shaving a smaller ground truth noise doesn't necessarily show that the normalization method is necessary better. But it's useful to see the amount each normalization cause as a reference line to compare other models. 
 %We use 1000 held-out point correspondences to evaluate the metrics described in Sec.~\ref{sec:evaluation}. 
 We compare our method with 8-point, LeMedS and RANSAC algorithms~\cite{zhang1998determining}. On average, $60$ pairs of keypoints are used per image. 
%Note that the ground truth values for training the network are computed using the RANSAC method. 
As we can observe, the reconstruction module is highly effective, and without it the network is unable to recover accurate fundamental matrices. The position features are also helpful in decreasing the error. The Siamese network outperforms the Single-Stream architecture, and can achieve errors comparable to the ground truth. {This shows that the two streams used to process each of the input images are indeed useful.} %Using the reconstruction module based on camera parameters slightly outperforms the epipolar parametrization. This might be due to the cases in which the first two columns of the F-matrix are linearly dependent.   
Note that the networks are trained end-to-end without the need for extracting point correspondences between the images, yet they are able to achieve competitive results with classic algorithms. The epipolar parametrization generally outperforms the other methods. During the inference time, we just need to pass the images to the feature extraction and regression networks to estimate the fundamental matrices.   

% The prediction outputted from the neural network is significantly better than ground truth generated from OpenCV in terms of \textbf{EPI-ABS}, \textbf{EPI-SQR} and \textbf{SED} metrics. However, the \textbf{SSD} (Sampson Distance) metric for the prediction is consistently higher compared to the ground truth.%; in the contrary, the neural network could outperform the OpenCV in \textbf{SED}. 
%  We are still investigating the source of this phenomenon.

% Adding the reconstruction module significantly reduces the error no matter which kind of feature extractor it is based on. This shows the effectiveness of the reconstruction module. But adding the position feature does not show consistent improvement. Our hypothesis is that since right now the stereo-images pairs are passed into the network by channels, the position features might not provide meaningful gain in this setting. A future work will be to use Siamese structure and extract position features separately from each image.

% It is worth noting that the large error in OpenCV ground truth can be caused by the artifact from the synthetic images. Usually, the metrics \textbf{EPI-SQR} should get a value of less than $1$ in order to be considered to be a good F-matrix. Therefore, another direction for future work will be to improve the quality of the dataset or run on other real-world datasets such as KITTI~\cite{geiger2013vision}. 

\section{Conclusion and Future Work}
We present novel deep neural networks for estimating fundamental matrices from a pair of stereo images. 
Our networks can be trained end-to-end without the need for extracting point correspondences. We consider two different network architectures for computing features from the images, and show that the best result is obtained when we first process images in two streams, and then concatenate the features and pass the result to a single-stream network.   
We show that the simple approach of directly regressing the nine entries of the fundamental matrix does not yield good results. Therefore, a reconstruction module is introduced as a differentiable layer to estimate the parameters of the fundamental matrix. Two different parametrizations of the F-matrix are considered: one based on the camera parameters, and the other based on the epipolar parametrization.  
We also demonstrate that position features can be used to further improve the estimation. 
This is due to the sensitivity of fundamental matrices to the location of points in the input images. 
In the future, we plan to extend the results to other datasets, and explore other parametrizations of the fundamental matrix. 

%
% ---- Bibliography ----
%
% BibTeX users should specify bibliography style 'splncs04'.
% References will then be sorted and formatted in the correct style.
%
% \bibliographystyle{splncs04}
% \bibliography{mybibliography}
%
\bibliographystyle{splncs04}
\bibliography{egbib}
% \begin{thebibliography}{8}
% \bibitem{ref_article1}
% Author, F.: Article title. Journal \textbf{2}(5), 99--110 (2016)

% \bibitem{ref_lncs1}
% Author, F., Author, S.: Title of a proceedings paper. In: Editor,
% F., Editor, S. (eds.) CONFERENCE 2016, LNCS, vol. 9999, pp. 1--13.
% Springer, Heidelberg (2016). \doi{10.10007/1234567890}

% \bibitem{ref_book1}
% Author, F., Author, S., Author, T.: Book title. 2nd edn. Publisher,
% Location (1999)

% \bibitem{ref_proc1}
% Author, A.-B.: Contribution title. In: 9th International Proceedings
% on Proceedings, pp. 1--2. Publisher, Location (2010)

% \bibitem{ref_url1}
% LNCS Homepage, \url{http://www.springer.com/lncs}. Last accessed 4
% Oct 2017
% \end{thebibliography}
\end{document}